\documentclass{ieeeaccess}
\usepackage{amsmath,amssymb,amsfonts}
\usepackage{algorithmic}
\usepackage{textcomp}
\usepackage{booktabs} 
\usepackage{multirow}
\usepackage[numbers]{natbib}
\usepackage{url}
\usepackage{float}
\usepackage[T1]{fontenc}
\usepackage[utf8]{inputenc}
\usepackage{times}
\usepackage{latexsym}
\usepackage{inconsolata}
\usepackage{microtype}
\usepackage{makecell}
\usepackage{graphicx}

\newcommand{\ours}{Cawai}
\newcommand{\Ours}{\textsc{Causality AWAre dense retrIever}}

\def\BibTeX{{\rm B\kern-.05em{\sc i\kern-.025em b}\kern-.08em
    T\kern-.1667em\lower.7ex\hbox{E}\kern-.125emX}}
\begin{document}
\history{Date of publication xxxx 00, 0000, date of current version xxxx 00, 0000.}
\doi{10.1109/ACCESS.2017.DOI}

\title{Causal Retrieval via Semantic Regularization}
\author{\uppercase{Hyunseo Shin}\authorrefmark{1},
\uppercase{Youngrok Choi\authorrefmark{2}, and Wonseok Hwang}.\authorrefmark{1},
\IEEEmembership{Member, IEEE}}
\address[1]{Department of Artificial Intelligence, University of Seoul, Seoul 02504, South Korea}
\address[2]{Naver Cloud, Seongnam-si 13561, South Korea  (e-mail: youngrok.choi@navercorp.com)}

\tfootnote{This work was supported by the National Research Foundation of Korea(NRF) grant funded by the Korea government (MSIT: Ministry of Science and Information and Communication Technology) (RS-2025-23524855) and by the 2024 Advanced Facility Fund of the University of Seoul.}

\markboth
{Author \headeretal: Preparation of Papers for IEEE TRANSACTIONS and JOURNALS}
{Author \headeretal: Preparation of Papers for IEEE TRANSACTIONS and JOURNALS}

\corresp{Corresponding author: Wonseok Hwang (e-mail: wonseok.hwang@uos.ac.kr).}

\begin{abstract}

To extend the capabilities of Large Language Models (LLMs) to knowledge-intensive domains, it becomes a standard practice to combine an information retrieval (IR) system with LLMs.
However, for IR systems to effectively support such applications, they must move beyond simple semantic matching and accurately capture diverse query intents, including causal relationships.
Existing IR models primarily focus on retrieving documents based on surface-level semantic similarity, often overlooking deeper relational structures such as causality.
To address these limitations, we propose \ours, a retriever trained with dual objectives that jointly capture semantic and causal relations via semantic-regularization mechanism.
Our extensive experiments demonstrate that \ours\ outperforms various models on diverse causal retrieval tasks, especially under large-scale retrieval settings. 
We also show that \ours\ exhibits strong zero-shot generalization across scientific domain QA tasks.
Finally, we find that when combined with a conventional semantic dense retriever, \ours\ achieves superior performance even on general QA tasks, highlighting its complementary strengths.


\end{abstract}


\titlepgskip=-15pt

\maketitle

\section{Introduction}

With recent advancements in large language models (LLMs), it has become standard practice to enhance the performance of LLMs via retrieval-augmented generation (RAG). In these systems, the retriever is responsible for identifying relevant documents; however, retrieval errors often cascade into generation errors, producing incorrect reasoning. Indeed, recent analysis in legal domains shows that 40–50\% of hallucinations originate in the retrieval stage, underscoring its critical importance \cite{magesh2024hallucinationfreeassessingreliabilityleading}.

Traditionally, information retrieval (IR) systems define relevance in terms of semantic similarity.
However, the notion of ``relevant'' often encompasses various aspects 
\citep{vanOpijnen2017ai_and_law_relevance}. 
While semantic relevance is suitable for tasks such as legal case retrieval, it is insufficient when user intent requires understanding causal relationships. This perspective is reinforced by ConvRAG~\citep{ye2024boostingconversationalquestionanswering}, which categorizes queries based on structure and intent, highlighting that retrieval requires a deeper alignment with the user’s informational needs.

In contrast, embedding-based retrieval tends to prioritize semantic similarity above all else. As a result, they often struggles to capture other forms of relevance—particularly causal relationships.

We observe this phenomenon in the e-CARE dataset, a dataset for causal reasoning \citep{e-care}. A widely adopted dense passage retrieval (DPR) method \citep{dpr} effectively retrieves the next causal sentence when the retrieval pool is small, but when Wikipedia sentences are added, it retrieves sentences based on semantic similarity rather than true causal relationships. 
For instance, given the query \textit{``An explosion of Sulfides occurred in the factory.''} the correct effect is \textit{``The workers were all injured due to  eye irritation to suffocation.''}. Yet, DPR selects \textit{``On 22 February 2003, one of the production facilities caught fire and was badly damaged.''}, semantically similar yet causally irrelevant passages. A manual analysis of 50 randomly sampled cases where DPR retrieved an incorrect passage 
reveals that 44\% of failures arise from such semantic drift. This suggests a limitation of existing retrieval models in causal tasks. Retrievers need a way to separate true causal relevance from spurious semantic association.

To address this challenge,
we propose \textbf{\ours}\footnote{\Ours}, a dense retriever that uses a semantic-regularization mechanism to disentangle causal signals from spurious semantic similarity.

\ours\ introduces dual constraints: (1) a \textbf{causal loss}, which encourages representations that reflect causal relations between query and document, and (2) a \textbf{regularization loss}, that anchors the representations to semantic features provided by a frozen encoder.
The regularization minimizes  semantic information loss during causal representation learning and can also be viewed as an indirect deconfounding mechanism within the causal-inference framework (Section \ref{sec:causal-interp}).
Collectively these two losses enable the model to avoid spurious semantic matches and focus on retrieving \textit{causally} relevant passages that are aligned with the query. 

Evaluation results show that \ours\ significantly outperforms existing baselines, such as BM25, DPR, GTR~\citep{ni-etal-2022-large}, and BGE-M3~\citep{bge_m3} in causal retrieval, causal QA, and scientific QA tasks, while achieving comparable performance on general QA tasks (Table 2--5). Furthermore, when \ours\ is integrated with a baseline retriever, the resulting hybrid system achieves superior performance even on general QA tasks (Table 5). These results indicate that while \ours\ provides most advantages on causal tasks, it also contributes orthogonal gains when combined with existing semantic-based retrieval systems.

In summary, our contributions are as follows.
\begin{itemize}
    \item We propose \ours, a dense retriever that is specialized in causal tasks.
    \item \ours\ achieves significantly better performance in causal tasks, and comparable performance in general QA tasks compared to existing dense retrieval baselines.
    \item When integrated with a conventional retriever, \ours\ also demonstrates advantages on general QA tasks, contributing orthogonal gains\footnote{The code is available at \url{https://github.com/00HS/causality-aware-retriever}}.
\end{itemize}

\section{Related Work}
\subsection{Information Retrieval}
Traditional keyword-based information retrieval methods like BM25 \citep{bm25} rely heavily on lexical overlap between queries and documents, which limits their ability to capture deeper semantic relationships.

To address this limitation, dense retrieval methods such as Dense Passage Retrieval (DPR) \citep{dpr} encode queries and documents into dense vectors using pretrained language models, enabling semantic matching.
More recent approaches leverage knowledge distillation to build compact yet effective embedding models Qwen3 Embedding \citep{qwen3embedding} and EmbeddingGemma \citep{vera2025embeddinggemmapowerfullightweighttext}.
Gecko \citep{lee2024geckoversatiletextembeddings} distills representations directly from large language models, while BGE-M3 \citep{bge_m3} distills from multiple task-specific embedding models. 




Recent research has explored architectural modifications and hybrid objective functions to further improve retrieval precision. For instance, \cite{modified} propose a modified cross-encoder that effectively distinguishes relevant documents by minimizing a combined loss function of sigmoid cross-entropy and cosine similarity.

Compared to these works, our work is different in that we focus on ``causal retrieval'' while introducing three specialized encoders for deconfounding.

\subsection{Causal Relationship Identification}
Recent works in causal discovery with LLMs focus on identifying cause-effect relationships by leveraging causal graphs.
A framework for causal discovery is introduced in Causal Parrots~\citep{causalparrotslargelanguage}, where LLMs can return causal graphs through conditional independence statements. 
Zhang et al. \cite{zhang2024causal} proposes LACR, a retrieval-augmented method that leverages LLMs to recover causal graphs from text. Using retrieved evidence, the method infers structural relationships among variables. Similarly, CausalRAG \cite{wang-etal-2025-causalrag} incorporates causal graphs into the retrieval process to preserve contextual continuity and improve precision beyond simple semantic matching.
 While these methods offer insights for post-hoc causal analysis, they apply causal reasoning only after retrieval. In contrast, our work incorporates causal cues directly into retrieval, enabling the model to identify causal relationships in the early stage.

\section{Method}
In this section, we present our causal retrieval model \ours,  beginning with the overall formulation and followed by the model architecture and training objectives. It is important to note that Cawai is designed as a causality-aware representation learning framework inspired by causal inference, rather than a formal causal inference system for estimating strict causal effects.

\subsection{\ours} 

\Figure[t!](topskip=0pt, botskip=0pt, midskip=0pt)[width=.95\textwidth]{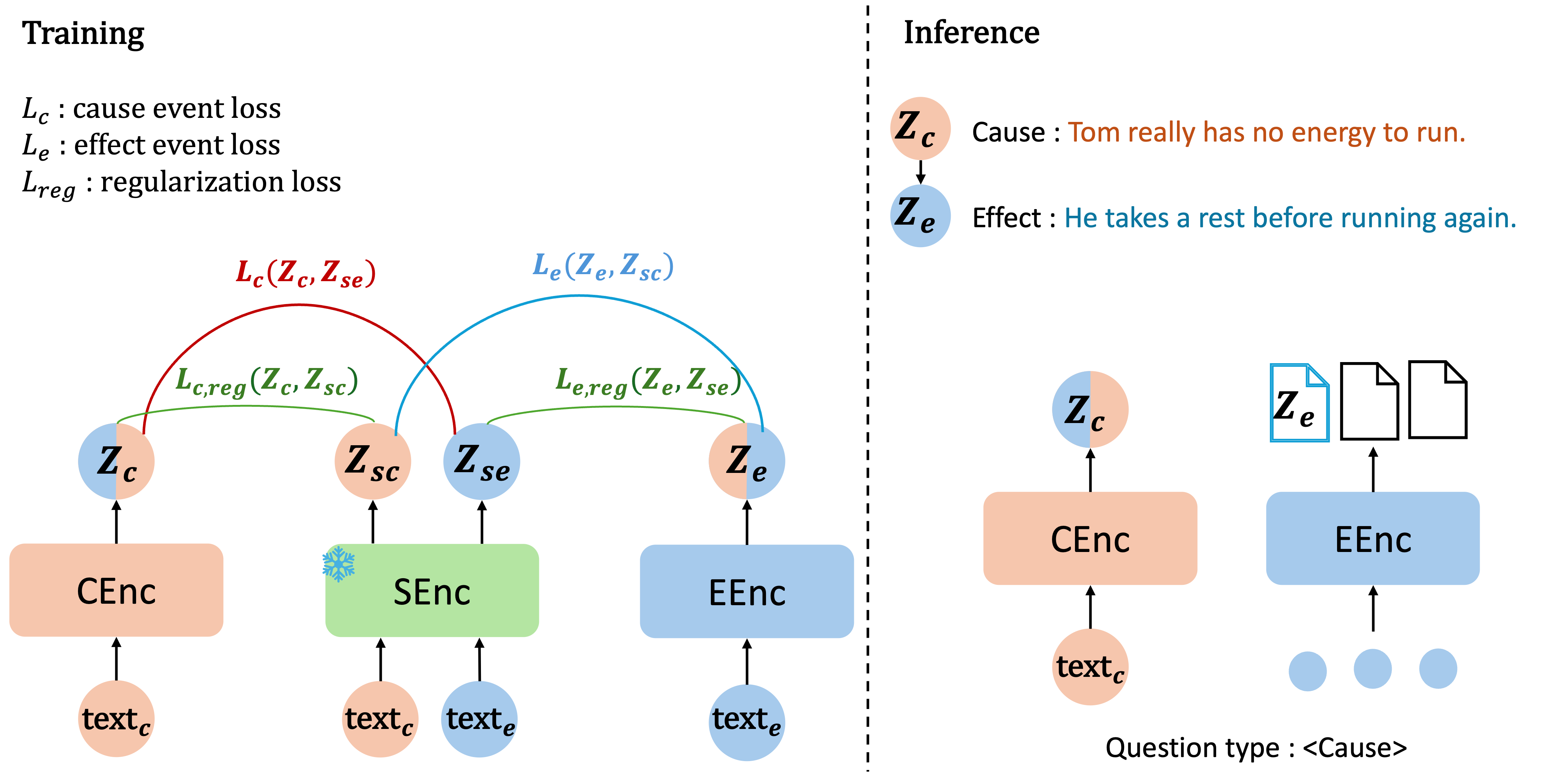}{Overall architecture of \ours. The framework consists of three encoders: CEnc, EEnc, and SEnc. CEnc encodes cause texts into effect representations ($c$), while EEnc encodes effect texts into cause representations ($e$). SEnc serves as a shared semantic baseline, aligning its outputs with those of both CEnc and EEnc. The red, blue, and green lines represent $L_c$, $L_e$, and $L_{\text{reg}}$, respectively. During inference, only CEnc and EEnc are used.\label{fig:causal_retrieval}}


\paragraph{Model Architecture}
\ours\ employs three encoders: CEnc, EEnc, and SEnc (Figure \ref{fig:causal_retrieval}). All three encoders share an identical Transformer-based architecture and are initialized with the same pre-trained weights from a backbone model (e.g., BERT, GTR, or BGE-M3). While CEnc and EEnc are optimized to capture causal relations, SEnc is kept frozen throughout the training process to provide a stable semantic reference.
CEnc encodes the cause event text (e.g. {\it Tom really has no energy to run.}), denoted as text$_c$, generating a vector representation $z_c$. Similarly, EEnc encodes the effect event text$_e$ (e.g. {\it He takes a rest before running again.}), producing an embedding $z_e$. 
SEnc, whose parameters remain frozen throughout training, processes both $\text{text}_c$ and $\text{text}_e$ independently and generates semantic vector representations $z_{sc}$ and $z_{se}$, respectively. These semantic vectors serve as regularization signals to minimize distortion in semantic representation during causal representation learning.

\paragraph{Training}
CEnc and EEnc are trained to map a cause text$_c$ and an effect text$_e$ into a representation space that captures causal relations. SEnc remains fixed during training and provides semantic information for regularization.

We use in-batch negative sampling across all three encoders.
For CEnc, given a pair $(\text{text}_{c,i}, \text{text}_{e,i})$, we treat the remaining effect texts
$\{\text{text}_{e,j} \mid j \neq i\}$ in the batch as negatives.  
The causal alignment loss is defined as:

\begin{equation}
\mathcal{L}_{c}(\text{text}_{c,i}, \text{text}_{e,i}) = - \log \frac{\exp(s(z_{c,i}, z_{se,i}))}{\sum_{j \in \text{batch}}\exp(s(z_{c,i}, z_{se,j}))},
\end{equation}
where $s(\cdot,\cdot)$ denotes cosine similarity.

 To minimize the distortion in semantic representation while learning causal relationship, we introduce a regularization loss that aligns the causal representation with the frozen semantic representation:

\begin{equation}
\begin{aligned}
\mathcal{L}_{c, \text{reg}}(\text{text}_{c,i}) &=
- \log \frac{\exp(s(z_{c,i}, z_{sc,i}))}{
\sum_{j \in \text{batch}}\exp(s(z_{c,i}, z_{sc,j}))}
\end{aligned}
\end{equation}

Likewise, EEnc is trained with symmetric causal and regularization losses:

\begin{equation}
\begin{aligned}
\mathcal{L}_{e}(\text{text}_{c,i}, \text{text}_{e,i}) &= - \log \frac{\exp(s(z_{e,i}, z_{sc,i}))}{\sum_{j \in \text{batch}} \exp(s(z_{e,i}, z_{sc,j}))} 
\\\mathcal{L}_{e, \text{reg}}(\text{text}_{e,i}) &=
- \log \frac{\exp(s(z_{e,i}, z_{se,i}))}{
\sum_{j \in \text{batch}}\exp(s(z_{e,i},z_{se,j}))}
\end{aligned}
\end{equation}


The total training objective is the sum of causal alignment losses and regularization losses:

\begin{equation} \label{eq:total_loss}
\mathcal{L}_{total} = 
 \mathcal{L}_{c} 
 + \mathcal{L}_{e} 
+ \beta(\mathcal{L}_{c, \text{reg}} + \mathcal{L}_{e, \text{reg}})
\end{equation}
where $\beta$ controls the weight of the regularization terms.





\paragraph{Inference}
During inference, only CEnc and EEnc are used.  
Given an input cause text, we compute retrieval scores as  
\[
P(\text{text}_e \mid \text{text}_c) \approx \text{softmax}(z_c \cdot z_e),
\] 
following the standard dense retriever. 
Since SEnc is no longer required at inference time, \ours\ shows the same inference-time efficiency as conventional dense retrievers.


\section{Experiments}

\subsection{Datasets}
\subsubsection{e-CARE  Evaluation}
We use the e-CARE \cite{e-care} and BCOPA-CE \citep{balanced-copa-ce} for causal retrieval experiments. Although these datasets were originally designed for causal reasoning, we adapt them into a retrieval setting in order to evaluate whether \ours\ can effectively identify causally relevant information from large collections of text. The e-CARE dataset is divided into training, validation, and test splits following a 6:1:1 ratio. BCOPA-CE is used only for training and validation, consisting of 500 triplets of \texttt{<cause, premise, effect>}. We convert each triplet into two cause–effect pairs, \texttt{(cause, premise)} and \texttt{(premise, effect)}, resulting in 1,000 pairs. To prevent data leakage, pairs originating from the same triplet are kept within the same split (Table~\ref{tab:data_distribution}).
 
For retrieval evaluation, we construct large-scale retrieval pools from Wikipedia\footnote{\url{https://huggingface.co/datasets/wikimedia/wikipedia}} and RedPajama-Data-v2 \citep{together2023redpajama}\footnote{\url{https://huggingface.co/datasets/togethercomputer/RedPajama-Data-V2}} to simulate real-world retrieval scenarios. We use varying retrieval pool sizes: ranging from 2 million sentences ($XL$) to 20 million sentences ($XXL$).

\subsubsection{Causal QA Evaluation}

To evaluate \ours\ on real-world causal question-answering tasks, we follow the retrieval environments defined in CausalQA \citep{bondarenko2022causalqa}.  We use four datasets: MS MARCO \cite{bajaj2018msmarcohumangenerated}, Natural Questions \cite{kwiatkowski-etal-2019-natural}, SQuAD v2.0 \cite{rajpurkar-etal-2018-know}, and HotpotQA \cite{yang-etal-2018-hotpotqa} for training, validation, and test (Table \ref{tab:data_distribution}).

\begin{table}[h]
    \centering
    \begin{tabular}{lccc}
        \toprule
        Dataset \textit{(\% of Total)} & Train & Validation & Test \\  
        \midrule
        e-CARE & 12,792 & 2,132 & 2,136 \\
        BCOPA-CE & 900 & 100 & -
        \\
        \hline
        HotpotQA \textit{(0.4\%)} & 312 & 39 & 39  \\  
        MS MARCO \textit{(2.5\%)} & 19,318 & 2,415 & 2,415  \\  
        SQuAD v2.0 \textit{(2.3\%)} & 2,567 & 321 & 321 \\  
        Natural Questions \textit{(0.4\%)} & 968 & 120 & 120 \\  
        \bottomrule
    \end{tabular}
    \caption{Distribution of causal questions across datasets.}
    \label{tab:data_distribution}
\end{table}


\subsection{Experimental Setups}
\subsubsection{Baseline}
We include BM25 \cite{bm25}, DPR \cite{dpr}, GTR \cite{ni-etal-2022-large}, BGE-M3 \cite{bge_m3}, and LLaMA-1.0B as our baselines.  
We train all baseline retrievers (DPR, GTR, BGE-M3, LLaMA-1.0B) and their corresponding \ours\ variants under identical training settings, using the same pretrained initialization for each model family: BERT-Base-Uncased\footnote{https://huggingface.co/google-bert/bert-base-uncased}, GTR-Base\footnote{https://huggingface.co/sentence-transformers/gtr-t5-base}, BGE-M3\footnote{https://huggingface.co/BAAI/bge-m3}, and LLaMA-1.0B\footnote{https://huggingface.co/knowledgator/Llama-encoder-1.0B}. 

We train all models for 500 epochs with in-batch negatives, using a batch size of 64, a learning rate of 1e-5, and the AdamW optimizer. For LLaMA-1.0B, we fine-tune the encoder using LoRA adapters \citep{hu2022lora}. We select the best checkpoint based on validation accuracy.  
All experiments are conducted on NVIDIA A6000 or RTX 6000 Ada GPUs.
For CausalQA experiments (Table~\ref{causalqa}), we used BGE-M3-unsupervised\footnote{https://huggingface.co/BAAI/bge-m3-unsupervised}, since the supervised version was already fine-tuned on these datasets.

 \subsubsection{Metrics}
For causal QA datasets, we report Hit@1, Hit@10, and Mean Reciprocal Rank at 10 (MRR@10).  
For scientific-domain QA tasks, we use Normalized Discounted Cumulative Gain (NDCG).  
For general-domain QA experiments (Section~\ref{sec: general-domain-qa}), we do not employ a reader model; instead, we apply a simple fuzzy-matching approach as a relaxed evaluation metric, focusing solely on retrieval accuracy.

\begin{table*}[!hbt]
\setlength{\tabcolsep}{1pt}
\renewcommand{\arraystretch}{0.75}
\centering
\begin{tabular}{l|ccc|ccc|ccc}
    \toprule
    \multirow{2}{*}{\centering\textbf{Model}} 
    & \multicolumn{3}{c}{\textbf{e-CARE}} 
    & \multicolumn{3}{c}{\textbf{e-CARE + $\text{wiki}_{XL}$}} 
    & \multicolumn{3}{c}{\textbf{e-CARE + $\text{RedPajama}_{XL}$}} \\
    \cmidrule(lr){2-4} \cmidrule(lr){5-7} \cmidrule(lr){8-10}
    & Hit@1 & Hit@10 & MRR@10 & Hit@1 & Hit@10 & MRR@10 & Hit@1 & Hit@10 & MRR@10 \\
    \midrule
    \multicolumn{10}{c}{\textbf{Task 1. Cause to Effect}} \\
    \midrule

    BM25 
    & 8.9 & 21.8 & 12.7 & 4.6 & 8.5 & 5.8 & 4.9 & 9.3 & 6.1 \\

    \hline 
    
    DPR
    & 36.3 & \textbf{66.0} & \textbf{45.5} 
    & 16.0 & 28.2 & 19.4 
    & 13.7 & 25.8 & 17.3 \\

    \ours-DPR
    & \textbf{36.8} & 63.7 & 45.1 
    & \textbf{20.4} & \textbf{32.1} & \textbf{23.7} 
    & \textbf{18.1} & \textbf{29.4} & \textbf{21.3} \\

    \hline 
    
    GTR
    & 42.2 & \textbf{71.8} & \textbf{51.9} 
    & 20.5 & \textbf{39.0} & 26.1 
    & 18.1 & \textbf{35.6} & 23.0 \\

    \ours-GTR
    & \textbf{43.2} & 70.0 & 51.2
    & \textbf{21.6} & 36.2 & \textbf{26.2} 
    & \textbf{19.6} & 34.6 & \textbf{24.1} \\

    \hline 
    
    BGE M3
    & 42.2 & 70.8 & 51.0
    & 22.1 & 40.1 & 27.5
    & 19.0 & 35.3 & 23.7 \\

    \ours-BGE M3
    & \textbf{46.5} & \textbf{73.0} & \textbf{54.5}
    & \textbf{32.1} & \textbf{50.0} & \textbf{37.3}
    & \textbf{28.9} & \textbf{45.4} & \textbf{33.7} \\

    \hline 

    LLaMA 1B (Lora)
    & 14.9 & 35.3 & 20.7 
    & 2.3 & 6.4 & 3.4 
    & 1.9 & 5.8 & 2.9 \\

    \ours-LLaMA 1B (Lora)
    & \textbf{23.5} & \textbf{51.0} & \textbf{31.6} 
    & \textbf{3.1} & \textbf{9.0} & \textbf{4.8} 
    & \textbf{3.6} & \textbf{9.5} & \textbf{5.1} \\

    \midrule

    \multicolumn{10}{c}{\textbf{Task 2. Effect to Cause}}   \\

    \midrule

    BM25 
    & 9.4 & 20.6 & 12.6 & 4.9 & 8.9 & 6.0 & 4.6 & 9.2 & 5.9 \\

    \hline

    DPR
    & \textbf{39.4} & \textbf{67.8} & \textbf{47.8} 
    & 17.3 & 32.4 & 21.9 
    & 14.4 & 28.8 & 18.6 \\

    \ours-DPR
    & 37.6 & 64.2 & 45.7 
    & \textbf{22.3} & \textbf{35.5} & \textbf{26.4} 
    & \textbf{20.9} & \textbf{34.0} & \textbf{24.8} \\

    \hline

    GTR
    & \textbf{42.8} & \textbf{72.4} & \textbf{52.2} 
    & \textbf{21.8} & \textbf{39.5} & \textbf{27.1} 
    & 18.3 & 36.0 & 23.5 \\

    \ours-GTR
    & 41.9 & 68.6 & 50.0 
    & 21.5 & 38.5 & 26.7 
    & \textbf{20.1} & \textbf{36.3} & \textbf{24.7} \\

    \hline

    BGE M3
    & 39.3 & 69.6 & 48.9
    & 21.1 & 39.7 & 26.8
    & 17.7 & 34.1 & 22.6 \\

    \ours-BGE M3
    & \textbf{46.3} & \textbf{72.9} & \textbf{54.6}
    & \textbf{33.6} & \textbf{49.2} & \textbf{38.5}
    & \textbf{30.3} & \textbf{45.3} & \textbf{34.8} \\

    \hline
    
    LLaMA 1B (Lora)
    & 15.1 & 35.4 & 20.8 
    & 3.3 & 9.7 & 5.1 
    & 3.0 & 8.9 & 4.8 \\

    \ours-LLaMA 1B (Lora)
    & \textbf{22.1} & \textbf{50.7} & \textbf{30.5} 
    & \textbf{4.4} & \textbf{12.3} & \textbf{6.5} 
    & \textbf{4.6} & \textbf{12.4} & \textbf{6.7} \\

    \bottomrule
\end{tabular}
\caption{Accuracy comparison on e-CARE. In Task 1, a model needs to retrieve the corresponding effect sentence for given cause text as a query whereas in Task 2, the model receive effect sentence and retrieve corresponding cause sentence.}
\label{performance_retrieval}
\end{table*}

\section{Results}
\subsection{Causal Retrieval Tasks}

We reformulate the e-CARE causal reasoning dataset into two causal retrieval tasks. In Task 1, the model retrieves the effect sentence given a cause query; in Task 2, it retrieves the cause sentence given an effect query. 
Under the small-pool setting (2,136 sentences), \ours\ performs comparable or stronger than baseline retrievers, as shown in Table~\ref{performance_retrieval} (cols 1–3 for both Task 1 and Task 2).




Next, we evaluate performance under larger retrieval pools of 2 million sentences (XL) sampled from English Wikipedia or RedPajama, which introduce substantial semantic distractors and better approximate real-world retrieval conditions.
Across these settings, \ours\ consistently outperforms baselines in most cases (cols 4-9). 

Although GTR attains slightly higher performance on Task 2 with the $\text{wiki}_{XL}$ pool (cols 4--6), its performance drops with the larger pool, 20 million sentences ($\text{wiki}_{XXL}$): 12.9\% Hit@1, 25.3\% Hit@10, and 16.6\% MRR@10, compared to \ours's 14.6\%, 26.0\%, and 17.8\%. 
A similar trend is observed for Task 1, \ours\ maintains a performance advantage (GTR: 12.1\%, 24.9\%, and 16.0\% vs \ours: 12.4\%, 25.3\%, and 16.4\%).
These results highlight the enhanced generalization capability of \ours\ across diverse settings.

Using BGE-M3 as the backbone further confirms this trend.
Although BGE-M3 already performs competitive performance (e.g., 22.1\% Hit@1 and 27.5\% MRR@10 on Task 1 with $\text{wiki}_{XL}$), \ours-BGE-M3 model achieves substantial improvements across all metrics and settings (rows 6,7). On Task 1 ($\text{wiki}_{XL}$), \ours-BGE-M3 achieves +10.0\% Hit@1, +9.9\% Hit@10, and +9.8\% MRR@10 over baselines (cols 4--6).

Results with LLaMA-1.0B further confirm the superiority of \ours\ in causal retrieval tasks (bottom two rows).

Overall, these findings support the effectiveness of our dual-objective training and show that \ours\ consistently improves causal retrieval performance across diverse encoders.

\subsection{CausalQA Tasks} \label{ssec:causalqa}

\begin{table*}[hbt!]
  \setlength{\tabcolsep}{1pt}
  \renewcommand{\arraystretch}{0.75}
  \centering
  \begin{tabular}{l|ccc|ccc|ccc|ccc}
    \toprule
    \multirow{2}{*}{\textbf{Model}} & \multicolumn{3}{c|}{\textbf{MS MARCO}} & \multicolumn{3}{c|}{\textbf{Natural Questions}} & \multicolumn{3}{c|}{\textbf{SQuAD v2.0}} & \multicolumn{3}{c}{\textbf{HotpotQA}} \\
    \cmidrule(lr){2-4} \cmidrule(lr){5-7} \cmidrule(lr){8-10} \cmidrule(lr){11-13}
    & Hit@1&Hit@10 & MRR@10 
    & Hit@1&Hit@10 & MRR@10 
    & Hit@1&Hit@10 & MRR@10 
    & Hit@1&Hit@10 & MRR@10 
    \\

    \midrule

    DPR  
    & 10.6  & 38.3 & 18.4 
    & 2.2  & 16.0 & 4.7 
    & 11.2  & 24.3 & 15.0 
    & \textbf{3.4} & 12.8 & 5.4 

 \\
    \ours-DPR
    &\textbf{11.7} & \textbf{40.5 } & \textbf{19.8}
    & \textbf{4.7} & \textbf{16.5 } & \textbf{7.9}
    & \textbf{13.8} & \textbf{29.3 } & \textbf{18.4}
    & \textbf{3.4} &\textbf{13.7 } &\textbf{5.9}
    \\

    \hline 
    GTR
    & \textbf{21.3} & \textbf{71.5 } &\textbf{33.7}
    & 7.5  & 27.0  & 13.8 
    & 23.7  & 44.4  & 30.3 
    & 9.4  &\textbf{35.9}  & 16.4 
 \\
   \ours-GTR
    & 20.6  & 59.4  & 32.2 
    & \textbf{11.3} & \textbf{28.9} & \textbf{16.2}
    & \textbf{27.1} & \textbf{48.6} & \textbf{34.1}
    & \textbf{12.8} & 33.3  & \textbf{19.2}
\\

    \hline
    BGE-M3
    & \textbf{18.1} & \textbf{55.0} & \textbf{28.6}
    & 5.8 & 30.6 & 13.1
    & 26.2 & \textbf{48.6}  & 34.1
    & 12.8 & 33.3  & 18.3
    \\
    
    \ours-BGE-M3
    & 15.3 & 48.4 & 24.9
    & \textbf{15.7} & \textbf{31.4} & \textbf{21.1}
    & \textbf{29.3} & 47.0 & \textbf{34.9}
    & \textbf{15.4} & \textbf{35.9}  & \textbf{22.8}
    \\
    \bottomrule
  \end{tabular}
  \caption{Accuracy comparison on CausalQA. Scores are averaged over three runs with different random seeds.}
\label{causalqa}
\end{table*}

In CausalQA tasks, \ours\ consistently outperforms the three baselines in most cases (Table~\ref{causalqa}). 
Notably, \ours-BGE-M3 achieves the largest gains on Natural Questions and SQuAD v2.0, 
surpassing BGE-M3 by 9.9\% and 2.9\% in Hit@1, respectively (rows~5--6, cols~4 and~7). 
In contrast, the performance gap on \textsc{MS MARCO} is smaller, 
which we attribute to the high lexical overlap between questions and answers 
(i.e., high ROUGE similarity). 
A detailed analysis is provided in Section~\ref{sec:analysis} and Figure~\ref{fig:rouge_causalqa}.

\subsection{Science Domain QA Tasks}

\setlength{\tabcolsep}{1pt}
\renewcommand{\arraystretch}{0.75}

\begin{table}[hbt!]
  \centering
  
  \begin{tabular}{l|cc|cc|cc|cc}
    \toprule
    \multirow{2}{*}{\textbf{Model}} & \multicolumn{2}{c|}{\textbf{NFCorpus}} & \multicolumn{2}{c|}{\textbf{SciDocs}} & \multicolumn{2}{c|}{\textbf{SciFact}} & \multicolumn{2}{c}{\textbf{SciQ}} \\
    \cmidrule(lr){2-3} \cmidrule(lr){4-5} \cmidrule(lr){6-7} \cmidrule(lr){8-9}
    & 5 &20 
    & 5 &20 
    & 5 &20  
    & 5 &20 

    \\

    \midrule
    DPR
    & 6.2 & 11.5
    & 10.1 & 16.2
    & 23.5 & 28.8
    & 50.0 & 55.2
    
 \\
    \ours-DPR
    & \textbf{8.7} & \textbf{13.0}
    & \textbf{10.7} & \textbf{18.0}
    & \textbf{24.5} & \textbf{29.8}
    & \textbf{62.5} & \textbf{66.3}

\\

    \hline 
    GTR
    & 6.1  & 13.6 
    & 20.8  & 28.8
    & 38.1  & 43.7
    & 71.7  & 74.5
    
 \\
   \ours-GTR
    & \textbf{6.6 } & \textbf{14.4 }
    & \textbf{21.3 } & \textbf{30.3 }
    & \textbf{44.0 } & \textbf{48.6 }
    & \textbf{83.8 } & \textbf{84.8 }
    \\

    \hline
    BGE-M3
    &\textbf{ 7.0} & 13.1
    & 17.7 & 26.1
    & \textbf{42.8} & \textbf{48.4}
    & \textbf{77.5} & \textbf{80.2}
    \\
    \ours-BGE-M3
    & 6.7 & \textbf{13.9}
    & \textbf{21.1} & \textbf{29.7}
    & 39.5 & 45.5
    & 75.5 & 78.2
    \\
    \bottomrule
  \end{tabular}
  \caption{Zero-shot Document Retrieval Performance on Science Domain QA (nDCG@k = 5, 20).  We report the mean over three different random seeds.}
  \label{tab:scienceqa}
\end{table}

Following MixGR~\citep{cai-etal-2024-mixgr}, we evaluate the zero-shot generalization capability of dense retrievers trained on CausalQA using four science QA datasets: NFCorpus \cite{boteva2016}, SciDocs~\cite{specter2020cohan}, SciFact~\cite{wadden-etal-2020-fact}, and SciQ~\cite{welbl2017crowdsourcingmultiplechoicescience}.
As shown in Table \ref{tab:scienceqa}, \ours\ consistently achieves higher nDCG scores in most of cases, demonstrating its strong generalization performance.

\subsection{General Domain QA tasks} \label{sec: general-domain-qa}
\begin{table}[hbt!]
\setlength{\tabcolsep}{1pt}
\renewcommand{\arraystretch}{0.75}
  \centering
  \begin{tabular}{l|ccc|ccc}
    \toprule
    \multirow{2}{*}{\textbf{Model}} & \multicolumn{3}{c|}{\textbf{Natural Questions}} & \multicolumn{3}{c}{\textbf{SQuAD v1.1}} \\
    \cmidrule(lr){2-4} \cmidrule(lr){5-7}
    & Hit@1 & Hit@20 & Hit@100  & Hit@1 &Hit@20 & Hit@100 \\
    \midrule
    DPR & 30.6 & 75.2 & 86.3 
    & 25.3 & 66.3 & 80.4\\
    \ours-DPR & 33.9 & 74.8 & 86.0 
    & 24.2 & 63.9 & 79.3 \\
    Hybrid & \textbf{37.0} & \textbf{78.3} & \textbf{87.8} 
    & \textbf{28.5} & \textbf{68.5} & \textbf{82.0} \\
    \hline
    GTR
    & \textbf{41.4} & 81.9 & 90.2
    & 37.8 & 77.8 & 88.0 \\
    \ours-GTR
    & 37.5 & 79.2 & 88.5
    & 36.2 & 76.2 & 86.8 \\
    Hybrid
    & \textbf{41.4} & \textbf{82.0} & \textbf{90.3} 
    & \textbf{39.6} & \textbf{78.6} & \textbf{88.5}\\
    \hline
    BGE
    & \textbf{35.3} & 80.2 & 89.8
    & 40.6 & 80.3 & 90.0 \\
    \ours-BGE
    & 31.6 & 78.8 & 88.8    
    & 35.0 & 77.1 & 88.4 \\
    Hybrid
    & \textbf{35.3} & \textbf{81.3} & \textbf{90.4}
    & \textbf{41.0} &\textbf{ 81.1} & \textbf{90.3}
    \\
    
    \bottomrule
  \end{tabular}
  \caption{Accuracy comparison on General QA. We report the mean over three different random seeds.}
    \label{tab:generalqa}

\end{table}

Finally, we evaluate \ours\ on general QA tasks using Natural Questions and SQuAD v1.1 \cite{rajpurkar-etal-2016-squad}. We trained \ours\ under the same conditions as DPR~\citep{dpr} for fair comparisons: using a batch size of 128, the December 2018 English Wikipedia dump as the retrieval pool.

While \ours-GTR and \ours-BGE-M3 underperform their respective baselines, \ours-DPR achieves performance comparable to DPR without a clear advantage (Table~\ref{tab:generalqa}).  
This suggests that \ours\ is particularly effective in CausalQA settings, where causal relevance diverges from semantic similarity.

As \ours\ maps text to causal space while regularizing semantic information, it is especially helpful when the query and the corresponding causally connected documents have no significant keyword overlap. 
Indeed, our analysis indicates that baseline models tend to retrieve passages that are semantically similar but causally irrelevant, as shown in Table \ref{example} (see also Section \ref{sec:analysis}). 

Based on this observation, we construct a hybrid retriever that combines the score from \ours\ ($\text{score}_{C}$) and that of the corresponding baseline retriever ($\text{score}_{B}$). The final hybrid score between a query $q$ and a passage $p$ as
\begin{align}
    \text{score}&_{\text{hybrid}}(q, p) =\alpha \cdot \text{normalize}(\text{score}_{C}(q, p)) \nonumber \\
    &+ (1 - \alpha) \cdot \text{normalize}(\text{score}_{B}(q, p)),
\end{align}
where $\alpha \in [0, 1]$ controls the weighting, and min-max normalization over the top 100 candidates ensures score comparability.
 
The results show that the hybrid system achieves superior performance (Table~\ref{tab:generalqa}, cols 3, 6, and 9), demonstrating that \ours\ contributes complementarily to existing retrievers when integrated.
 The parameter $\alpha$ is set to 0.5, based on hyperparameter search (Table~\ref{tab:alpha_generalqa} in Appendix). 

\subsection{Ablation study}

\begin{table*}[!hbt]
\setlength{\tabcolsep}{2pt}

\centering
\resizebox{\textwidth}{!}{
\begin{tabular}{c|p{22cm}}
\toprule

\textbf{Input} & \multicolumn{1}{c}{ "Why are clouds flat at the bottom?"} \\
\midrule
\multirow{3}{*}{DPR} 
&Top 1  : “sometimes clouds aren't flat on the bottom, although they usually are." \textcolor{red}{(Wrong)}\\
&Top 2 : “before we look at why clouds are flat on the bottom, a little review on how clouds form, or exactly what a cloud is, might be helpful." \textcolor{red}{(Wrong)}\\
&Top 3 : ”The cloud cover shape has different looks depending upon how many oktas (eighths of the sky) are covered by cloud." \textcolor{red}{(Wrong)} \\
\midrule

\multirow{4}{*}{\ours-DPR} 
&Top 1  : "clouds are flat on the bottom because this is the transition point where the temperature and pressure are at a point where the air cannot hold all of it's water in its gaseous form (clouds are made of water in its liquid or solid state)."   \textcolor{blue}{(Correct)} \\
&Top 2 : "before we look at why clouds are flat on the bottom, a little review on how clouds form, or exactly what a cloud is, might be helpful." \textcolor{red}{(Wrong)} \\
&Top 3 : "well, not all clouds are flat at the bottom, it is not good to assume that." \textcolor{red}{(Wrong)}
 \\

\bottomrule
\end{tabular}
}
\caption{Example of Model Responses comparison on CausalQA. "Correct" indicates that the retrieved passage correctly answers the question, while "Wrong" denotes an incorrect passage. Results related to BGE are provided in the Appendix \ref{example_bge}.}
\label{example}
\end{table*}

\paragraph{Effect of Regularization Loss}
\begingroup
\setlength{\tabcolsep}{0.5pt}
\renewcommand{\arraystretch}{0.75}

\begin{table}[!hbt]
\centering
  \begin{tabular}{l|ccc|ccc}
    \toprule
    \multirow{3}{*}{\textbf{Loss}} 

        & \multicolumn{6}{c}{\textbf{Task 1. Cause to Effect}} 

 \\
    \cmidrule(lr){2-7} 
    & \multicolumn{3}{c|}{\textbf{e-CARE}} 
    & \multicolumn{3}{c}{\textbf{e-CARE + $\text{wiki}_{XL}$}}  \\
    \cmidrule(lr){2-4} \cmidrule(lr){5-7} 
     & H@1 & H@10 & M@10 & H@1 & H@10 & M@10 \\
    \midrule
    $\mathcal{L}_{c}+\mathcal{L}_{e}$
    & 32.4 & 60.4 & 40.9 
    & 11.5 & 24.5 & 15.3  \\
    
    $\mathcal{L}_{c}+\mathcal{L}_{e} + 0.1~\mathcal{L}_{\text{reg}}$ 
    & 37.2 & 63.6 & 45.3 
    & 20.4 & 32.5 & 24.0  \\
   
    $\mathcal{L}_{c}+\mathcal{L}_{e} + 1~\mathcal{L}_{\text{reg}}$ 
    & 36.8 & 63.7 & 45.1
    & 20.4 & 32.1 & 23.7 \\
    
    $\mathcal{L}_{c}+\mathcal{L}_{e} + 2~\mathcal{L}_{\text{reg}}$ 
    & 38.3 & 64.3 & 46.1 
    & 22.8 & 33.9 & 26.0 \\
    
    $\mathcal{L}_{c}+\mathcal{L}_{e} + 5~\mathcal{L}_{\text{reg}}$ 
    & 38.5 & 64.7 & 46.3 
    & 22.3 & 34.1 & 25.8  \\

    \bottomrule
  \end{tabular}
    \caption{Accuracy comparison on regularization loss hyperparameters on e-CARE. 
  $\mathcal{L}_{\text{reg}}$, $H@k$, and $M@k$ stands for the sum of two regularization losses defined in Eqn. \ref{eq:total_loss}, Hit@k, and MRR@k respectively.
  }
    \label{semantic_loss}
\end{table}
\endgroup

\begin{table}[!hbt]
\setlength{\tabcolsep}{1pt}
\renewcommand{\arraystretch}{0.5}
\centering
\begin{tabular}{l|ccc|ccc}
    \toprule
    \multirow{3}{*}{\centering\textbf{Loss type}} 
    & \multicolumn{6}{c}{\textbf{Task 1. Cause to Effect}}  \\
    \cmidrule(lr){2-7} 
    & \multicolumn{3}{c}{\textbf{e-CARE}} 
    & \multicolumn{3}{c}{\textbf{e-CARE + $\text{wiki}_{XL}$}}  

  \\
    \cmidrule(lr){2-4} \cmidrule(lr){5-7} 
    & H@1 & H@10 & M@10 
    & H@1 & H@10 & M@10\\
    \midrule

    w/o $\mathcal{L}_{e, \text{reg}}$
    & -1.3 & -1.3 & -1.6 
    &-4.2 &-4.3 &-3.9

     \\
    \hline 
 
    w/o $\mathcal{L}_{\text{e}}$
    & -8.1 & -8.5 &-8.2 
    & -15.3 &-18.9 &-16.4
 \\

     
 
\bottomrule
   
\end{tabular}
\caption{Ablation study showing the performance change ($\Delta$) resulting from loss terms in \ours-DPR on e-CARE.}
\label{appendix_bidirectional_task1}
\end{table}

Next, we examine how the performance of \ours\ depends on the regularization loss under varying weights ($\beta$) of the regularization loss in Eqn. \ref{eq:total_loss}.
Integrating the regularization loss results in a noticeable improvement in accuracy (Table \ref{semantic_loss} row 1, $\beta=0$ vs. row 2, $\beta=0.1$), highlighting its importance. 
Beyond $\beta=1$, no clear improvement were observed (row 3--5), and thus we set $\beta=1$ in all other experiments. A similar trend is observed in Task 2 (Table~\ref{appendix_semantic_loss} in Appendix).

\paragraph{Effect of Bidirectional Loss}
To understand the importance of bidirectional loss in \ours, we perform two types of ablation experiments on Task 1: (1) remove the loss for $\mathcal{L}_{e, \text{reg}}$ (Table \ref{appendix_bidirectional_task1}, row 1), or (2) remove EEnc (row 2). In this case, the similarity between $z_c$ and  $z_{se}$ is used for inference.
Results show that removing bidirectional regularization on either the query (cause) or document (effect) side significantly reduces performance, indicating that causal relationships in retrieval tasks benefit from symmetric regularization.
Additional experiments on Task 2 shows similar result (Table \ref{appendix_bidirectional_task2} in Appendix.)


\section{Analysis} \label{sec:analysis}

\Figure[!hbt](topskip=0pt, botskip=0pt, midskip=0pt)[width=0.95\columnwidth]{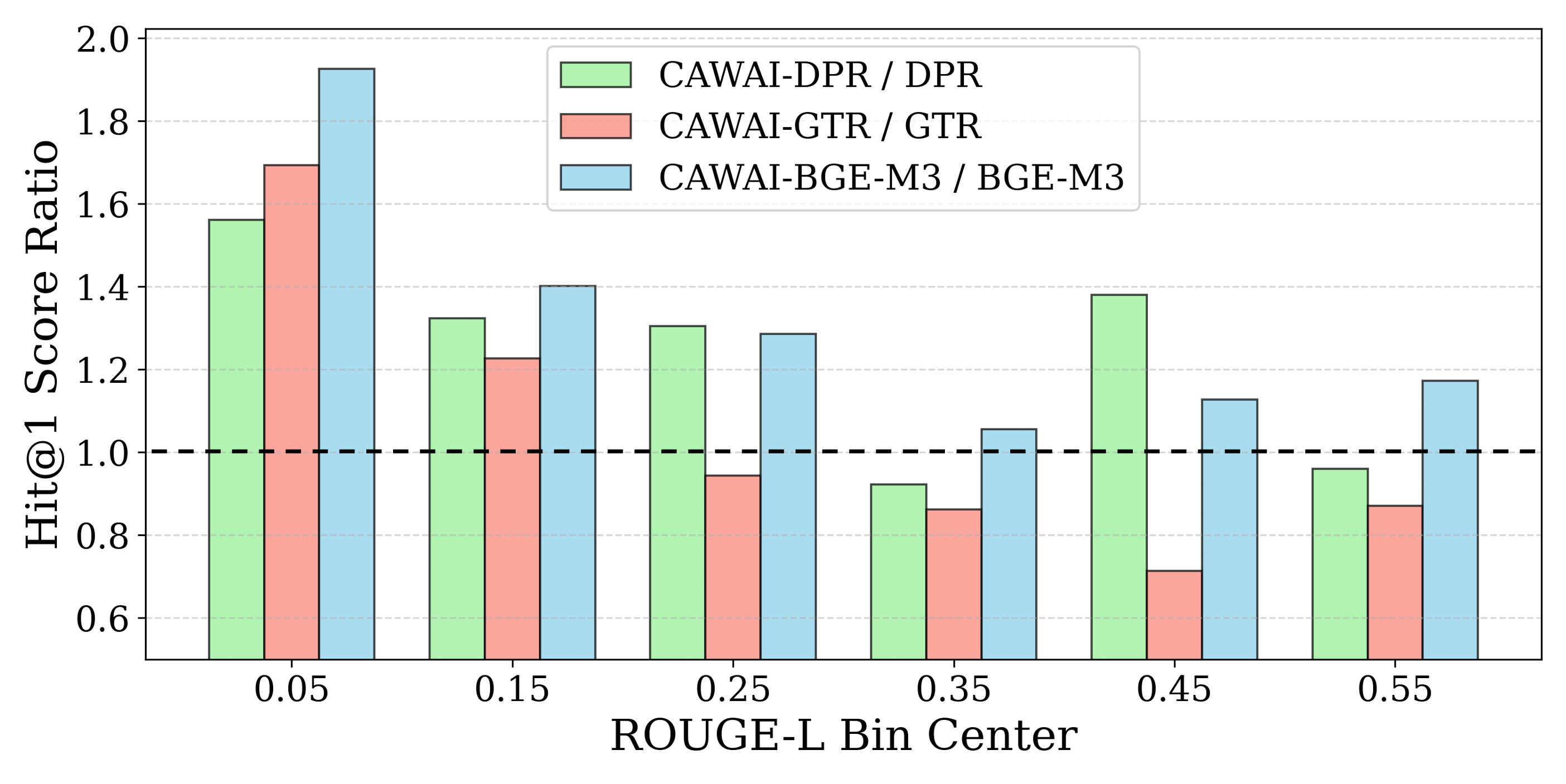}
{Hit@1 retrieval performance comparison on Task 1 of $\text{e-CARE}_{XL}$ (Table~\ref{performance_retrieval}). Hit@1 score ratio is computed as the \ours\ score divided by each baseline's score.\label{fig:rouge_e_care}}

\Figure[!hbt](topskip=0pt, botskip=0pt, midskip=0pt)[width=0.95\columnwidth]{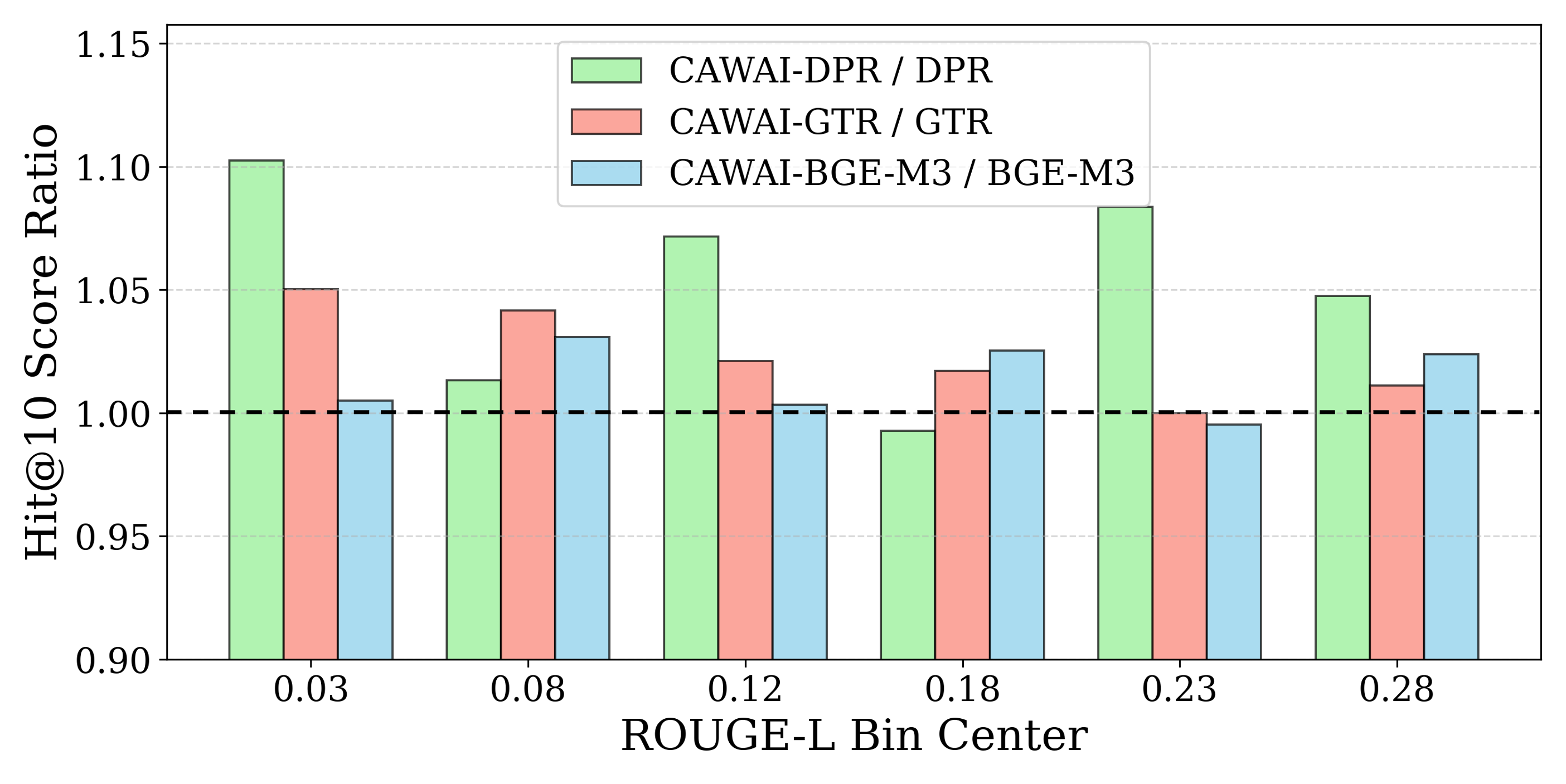}
{Hit@10 retrieval performance comparison on MS MARCO dataset from CausalQA. Hit@10 is used instead of Hit@1 due to the lower overall accuracy on MS MARCO (Table~\ref{causalqa}) compared to e-CARE (Table~\ref{performance_retrieval}).\label{fig:rouge_causalqa}}




To quantitatively analyze why \ours \ performs well in both causal tasks and general QA tasks (with hybrid retrieval), we plot the Hit@1 score ratio (y-axis) between \ours\ and baseline models as a function of query-document lexical overlap, measured by the ROUGE-L score (x-axis). Due to the limited number of test set size in the CausalQA datasets, this analysis focuses on the e-CARE (2,136 examples) and MS MARCO (2,415 examples) test sets for statistical significance (Table \ref{tab:data_distribution}).

As shown in Figures \ref{fig:rouge_e_care} and \ref{fig:rouge_causalqa}, \ours\ consistently outperforms baseline models when the ROUGE-L score is low showing the score ratio (=$\frac{\text{\ours}}{\text{Base Model}}$) $>$ 1 (see Table~\ref{tab:rouge_examples} for dataset examples and Table~\ref{appendix:rouge_bins_e_care}, Table~\ref{appendix:rouge_bins_msmarco} for ROUGE bin statistics).

These findings suggest that although Table \ref{causalqa} indicates \ours\ underperforms on MS MARCO in aggregate, it may in fact be better suited for causal retrieval tasks where lexical overlap between the query and target documents is minimal.


\section{Interpretation}
\label{sec:causal-interp}

The experimental results above reveal 
\ours\ achieves better performance on various causal tasks where semantic cues are weak or misleading.
How does \ours\ achieve such performance? To provide a conceptual interpretation of how our model works, we employ an interpretive framework from  causal inference. 






\subsection{Causal View of Relevance}

\begin{figure}
    \centering
    \includegraphics[width=0.8\linewidth]{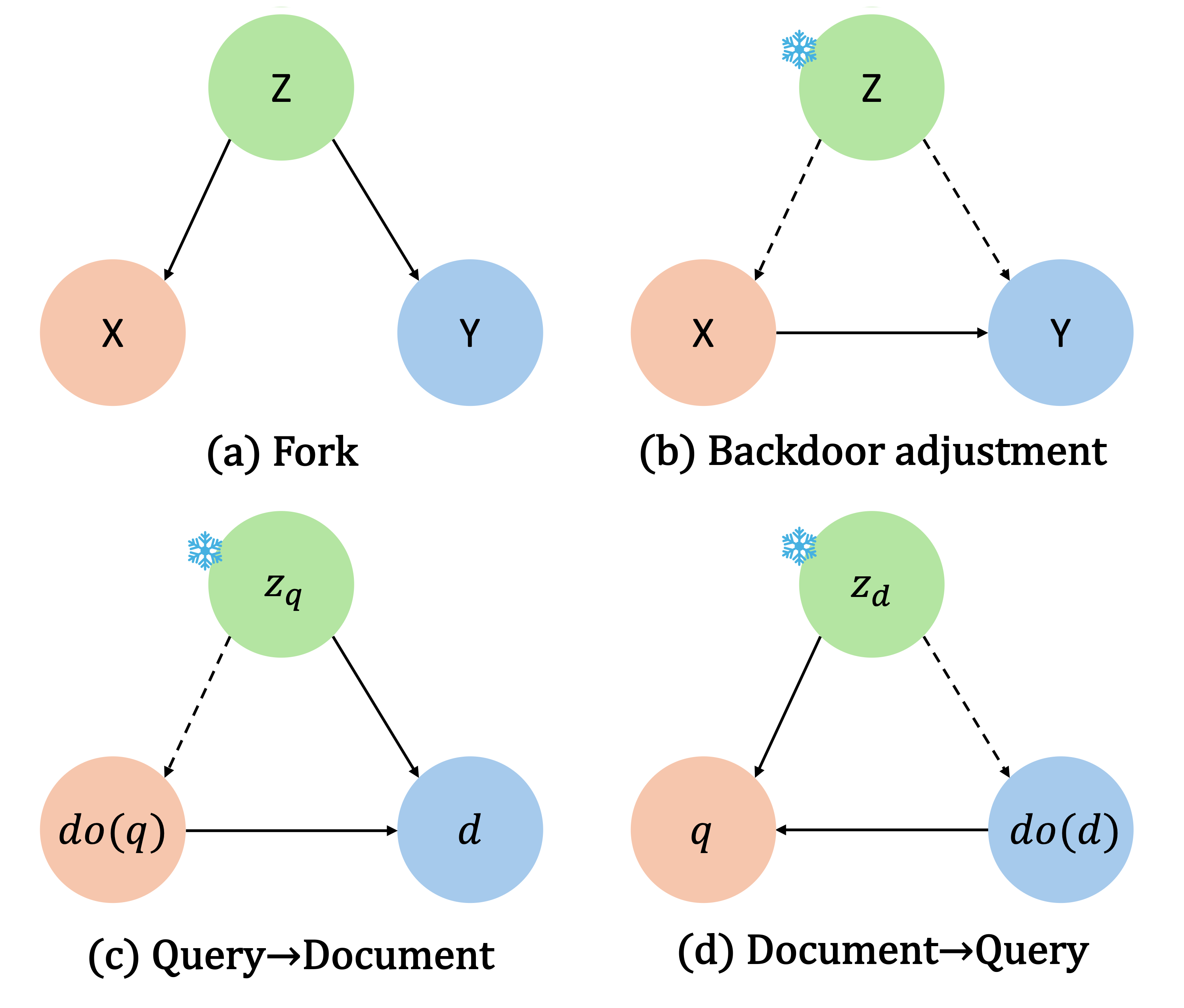}
    \caption{
    Illustrations of causal graphs in \ours.
(\textbf{a}) In the fork structure, both the query ($X$) and document ($Y$) are affected by a shared confounder ($Z$) such as semantic similarity, leading to spurious correlations.
(\textbf{b}) In backdoor adjustment, we condition on $Z$ (denoted by the snowflake icon), effectively blocking the backdoor path from $X$ to $Y$ and isolating the true causal effect on $X$ and $Y$.
(\textbf{c}) In the \textit{query $\rightarrow$ document} direction, $do(q)$ denotes an intervention where the query’s semantics ($z_q$) are used for deconfounding.
(\textbf{d}) In the reverse \textit{document $\rightarrow$ query} direction, the document’s semantics ($z_d$) serve as the confounder.}
  \label{fig:causal_graph}
\end{figure}

\paragraph{Relevance from Traditional IR}
In traditional IR, a document ($d$) is considered relevant to a query ($q$) if it conveys a similar meaning or addresses the same topic. 
This reflects a correlational view of relevance, as similarity metrics such as cosine similarity are closely related to statistical correlation \cite{altman2015points}  (Appendix\ref{proof_corr}).

\paragraph{Causal Relevance}

To define the notion of causal relevance, we leverage the concept of \textit{d-separation} from causal inference to isolate the true causal effect between query and document representations. 
D-separation is a key principle that tells us whether a set of variables can block all indirect paths—especially spurious paths—between two other variables in a causal graph~\citep{pearl2009causal,peter2017_elements_of_causal_inference}. 

We assume that semantic similarity between query and documents (the relevance from traditional IR) can create spurious causal pathways and if such backdoor paths are blocked, we can more reliably estimate the true causal effect between the variables of interest.

These concepts are demonstrated in Figure \ref{fig:causal_graph}. Consider a fork structure graph with three variables: $Z$, $X$, and $Y$ (a). We are interested in understanding the causal effect of variable $X$ (e.g., a query) on variable $Y$ (e.g., a document). However, both $X$ and $Y$ are influenced by a third variable $Z$ (e.g. topic, semantic): $Z \rightarrow X$ and $Z \rightarrow Y$. This results in a spurious association between $X$ and $Y$ because they both depend on $Z$, even if $X$ doesn’t actually cause $Y$ directly. If we ignore $Z$, we may wrongly attribute the effect of $Z$ 
to $X$, which can result in biased estimates of the causal effect from $X$ to $Y$.

To address this issue, we apply the backdoor adjustment technique to block the backdoor path $X \leftarrow Z \rightarrow Y$ by conditioning on $Z$ (Figure \ref{fig:causal_graph} (b)). This satisfies the d-separation condition, effectively removing the influence of the confounding variable $Z$ from our estimate. 

Reflecting this, we define the notion of causal relevance as below following Pearl~\citep{pearl2009causal}.

\begin{equation}
\begin{aligned}
P(D \mid do(q)) &\equiv \sum_z P(D \mid q, z) P(z),
\\P(Q \mid do(d)) &\equiv \sum_z P(Q \mid d, z) P(z),
\end{aligned}
\label{eq:backdoor-adjustment}
\end{equation}

where $do(q)$ denotes intervening on the query $q$ while holding the semantic confounder $Z$ fixed; similarly, $do(d)$ intervenes on the document $d$. 

This formulation is illustrated in the structure shown in Figure~\ref{fig:causal_graph}(c), where the query causally influences the document, and the query's semantic representation $z_q$ serves as the confounder to adjust for. In the reverse direction (Figure~\ref{fig:causal_graph}(d)), the document's semantics $z_d$ are used for deconfounding.


\subsection{Causal Dense Retriever}
\paragraph{Applying Backdoor Adjustment}
Standard dense retrievers optimize similarity between $q$ and $d$ directly. 
This leaves the backdoor path $q \leftarrow Z \rightarrow d$ unblocked, meaning semantic factors dominate relevance.



Rather than explicitly computing interventional probabilities, \ours\ implicitly approximates backdoor adjustment through its training objective.  
The frozen semantic encoder (SEnc) provides the confounder representation $z_{sc}, z_{se}$, and the regularization losses align each causal representation $(z_c, z_e)$ with its corresponding semantic confounder.  
This alignment encourages the model to condition on the confounder, effectively blocking the backdoor path.

\section{Conclusion}
We propose a novel retrieval method, \ours, that integrates cause-effect relationships into the retrieval process. Experiments show that \ours\ outperforms strong baselines in real-world causal retrieval, causal QA, and scientific QA tasks, while achieving comparable performance on general QA tasks. Furthermore, combining \ours\ with a conventional semantic retriever enhances general QA performance, indicating that it can provide orthogonal contributions to existing semantic retrievers.

\section{Limitations}
Our work aims to introduce dense retriever specialized in causal relationship. As a result, while it shows promising performance in causal QA tasks, it does not yield significant improvements in general QA settings without hybridization with conventional semantic retrievers.
Second, since the task focuses on retrieval rather than explicit reasoning, it differs from traditional causal inference settings that involve estimating intermediate (mediator) probabilities. 
While we employ a concept from causal inference framework to provide an interpretation regarding the role of the semantic regularization term, our explanation does not fully align with formal causal inference frameworks.
Third, the performance of Cawai is inherently dependent on the quality of the frozen semantic encoder (SEnc) used for regularization; any limitations in the backbone encoder’s semantic representation could potentially impact the effectiveness of the causal disentanglement. Furthermore, our current study is primarily focused on English-based retrieval. The generalization of causal retrieval to non-English languages or multi-modal scenarios, where causal cues may be represented through different modalities or linguistic structures, remains an unresolved challenge.

\appendices

\section*{Appendix}

\subsection{Effect of Regularization Loss}
\begin{table}[H]
\centering
  \begin{tabular}{l|ccc|ccc}
    \toprule
    \multirow{3}{*}{\textbf{Loss}} 

        & \multicolumn{6}{c}{\textbf{Task 2. Effect to Cause}} 

 \\
    \cmidrule(lr){2-7} 
    & \multicolumn{3}{c|}{\textbf{e-CARE}} 
    & \multicolumn{3}{c}{\textbf{e-CARE + $\text{wiki}_{XL}$}}  \\
    \cmidrule(lr){2-4} \cmidrule(lr){5-7} 
     & H@1 & H@10 & M@10 & H@1 & H@10 & M@10 \\
    \midrule
    $\mathcal{L}_{c}+\mathcal{L}_{e}$
    & 33.4 & 61.3 & 41.8 
    & 14.5 & 28.0 & 18.4 \\
    
    $\mathcal{L}_{c}+\mathcal{L}_{e} + 0.1~\mathcal{L}_{\text{reg}}$ 
    & 38.7 & 63.6 & 46.1 
    & 22.2 & 34.6 & 25.9 \\
   
    $\mathcal{L}_{c}+\mathcal{L}_{e} + 1~\mathcal{L}_{\text{reg}}$ 
    & 37.6 & 64.2 & 45.7 
    & 22.3 & 35.5 & 26.4 \\
    
    $\mathcal{L}_{c}+\mathcal{L}_{e} + 2~\mathcal{L}_{\text{reg}}$ 
    & 38.8 & 64.3 & 46.4 
    & 23.0 & 35.6 & 26.8 \\
    
    $\mathcal{L}_{c}+\mathcal{L}_{e} + 5~\mathcal{L}_{\text{reg}}$ 
    & 39.1 & 64.5 & 46.6 
    & 24.1 & 36.5 & 27.6 \\

    \bottomrule
  \end{tabular}
    \caption{Accuracy comparison on regularization loss hyperparameters on e-CARE.  
  $\mathcal{L}_{\text{reg}}$, $H@k$, and $M@k$ stands for the sume of two regularization losses defined in Eqn. \ref{eq:total_loss}, Hit@k, and MRR@k respectively.
  }
    \label{appendix_semantic_loss}
\end{table}

\subsection{Effect of Bidirectional Loss}
\begin{table}[H]
\setlength{\tabcolsep}{1pt}
\renewcommand{\arraystretch}{0.5}
\centering
\begin{tabular}{l|ccc|ccc}
    \toprule
    \multirow{3}{*}{\centering\textbf{Loss type}} 
    & \multicolumn{6}{c}{\textbf{Task 2. Effect to Cause}}  \\
    \cmidrule(lr){2-7} 
    & \multicolumn{3}{c}{\textbf{e-CARE}} 
    & \multicolumn{3}{c}{\textbf{e-CARE + $\text{wiki}_{XL}$}}  

  \\
    \cmidrule(lr){2-4} \cmidrule(lr){5-7} 
    & H@1 & H@10 & M@10 
    & H@1 & H@10 & M@10\\
    \midrule

    w/o $\mathcal{L}_{c, \text{reg}}$
    &-1.5 &-2.5 &-1.7 
    &-3.8 &-1.2 &-2.8
     
    \\
    \hline 
     w/o $\mathcal{L}_{\text{c}}$
    &-8.6 &-7.6 & -8.6
    &-15.1 & -19.8 & -16.4
    
    \\
 
\bottomrule
   
\end{tabular}
\caption{Ablation study on the loss terms in \ours-DPR on e-CARE. }
\label{appendix_bidirectional_task2}
\end{table}

\subsection{CausalQA}
The model responses for CausalQA are shown in Table~\ref{example_bge} below.

\begin{table*}[t]
\setlength{\tabcolsep}{2pt}

\centering
\resizebox{\textwidth}{!}{
\begin{tabular}{c|p{22cm}}
\toprule

\textbf{Input} & \multicolumn{1}{c}{ "Why are clouds flat at the bottom?"} \\
\midrule

\multirow{3}{*}{BGE-M3} 
&Top 1  :  "before we look at why clouds are flat on the bottom, a little review on how clouds form, or exactly what a cloud is, might be helpful." \textcolor{red}{(Wrong)} \\
&Top 2 : "sometimes clouds aren't flat on the bottom, although they usually are." \textcolor{red}{(Wrong)} \\
&Top 3 : "well, not all clouds are flat at the bottom, it is not good to assume that." \textcolor{red}{(Wrong)} \\
\midrule
\multirow{4}{*}{\ours-BGE-M3} 
&Top 1  : "clouds are flat on the bottom because this is the transition point where the temperature and pressure are at a point where the air cannot hold all of it's water in its gaseous form (clouds are made of water in its liquid or solid state)."  \textcolor{blue}{(Correct)}\\
&Top 2 : "this was the largest rebellion against soviet rule in azerbaijan in the 20th century, and caused the most losses."  \textcolor{red}{(Wrong)}\\
&Top 3 : "before we look at why clouds are flat on the bottom, a little review on how clouds form, or exactly what a cloud is, might be helpful." \textcolor{red}{(Wrong)} \\

\bottomrule
\end{tabular}
}
\caption{Example of Model Responses comparison on CausalQA. "Correct" indicates that the retrieved passage correctly answers the question, while "Wrong" denotes an incorrect passage.}
\label{example_bge}
\end{table*}

\subsection{GeneralQA}
\subsubsection{Hard Negatives Experiment}
\begin{table}[H]
\setlength{\tabcolsep}{1pt}
\renewcommand{\arraystretch}{0.85}
  \centering
  \begin{tabular}{l|ccc|ccc}
    \toprule
    \multirow{2}{*}{\textbf{Model}} & \multicolumn{3}{c|}{\textbf{Natural Questions}} & \multicolumn{3}{c}{\textbf{SQuAD v1.1}} \\
    \cmidrule(lr){2-4} \cmidrule(lr){5-7}
    & Hit@1 & Hit@20 & Hit@100  & Hit@1 &Hit@20 & Hit@100 \\
    \midrule

    DPR+BM25  & 35.1 & 74.7 & \textbf{86.7} & \textbf{27.1} & \textbf{69.4} & \textbf{83.7} \\
    \ours-DPR+BM25  &\textbf{36.6} & \textbf{75.9} & 86.1 & 25.1 & 65.9 & 81.2 \\

    \hline
    GTR+BM25 
    & \textbf{43.9} & \textbf{83.2} &\textbf{ 90.7}
    & \textbf{38.6} & \textbf{79.1} & \textbf{89.1} \\
    \ours-GTR+BM25 
     & 40.3 & 81.4 & 89.8
    & 36.7 & 76.4 & 87.2 \\

    \bottomrule
  \end{tabular}
  \caption{Accuracy comparison on General QA with a single BM25 hard negative}
    \label{tab:generalqa_bm25_hard_negative}
\end{table}

\subsubsection{Hybrid Experiment}
\begin{table}[H]
\setlength{\tabcolsep}{2pt}
\renewcommand{\arraystretch}{0.9}
\centering
\begin{tabular}{l|ccc|ccc}
\toprule
\multirow{2}{*}{\textbf{$\alpha$}} & \multicolumn{3}{c|}{\textbf{Natural Questions}} & \multicolumn{3}{c}{\textbf{SQuAD v1.1}} \\
\cmidrule(lr){2-4} \cmidrule(lr){5-7}
& Hit@1 & Hit@20 & Hit@100  & Hit@1 & Hit@20 & Hit@100 \\
\midrule

\multicolumn{7}{c}{$\alpha \ours + (1-\alpha) \text{DPR}$} \\
\midrule
0.1  & 31.9 & 75.7 & 87.5 & 26.3 & 67.0 & 81.4 \\
0.3  & 35.1 & 77.0 & 87.8 & 27.8 & 67.9 & 81.9 \\
0.5  & \textbf{37.0} & \textbf{78.3} & \textbf{87.8} & \textbf{28.5} & \textbf{68.5} & \textbf{82.0} \\
0.7  & 36.6 & 77.1 & 87.8 & 27.0 & 66.6 & 82.0 \\
0.9  & 35.0 & 75.6 & 87.4 & 25.2 & 64.8 & 81.2 \\
\midrule

\multicolumn{7}{c}{$\alpha \ours + (1-\alpha) \text{GTR}$} \\
\midrule
0.1  & 41.0 & 81.9 & 90.2 & 38.1 & 78.2 & 88.6 \\
0.3  & \textbf{41.4} & \textbf{82.0} & \textbf{90.3} & 39.0 & 78.8 & 88.6 \\
0.5  & 41.2 & 81.5 & 90.0 & \textbf{39.6} & \textbf{78.6} & \textbf{88.5} \\
0.7  & 40.0 & 80.8 & 89.9 & 39.5 & 77.6 & 88.3 \\
0.9  & 38.9 & 79.6 & 89.5 & 38.6 & 76.8 & 88.1 \\
\midrule

\multicolumn{7}{c}{$\alpha \ours + (1-\alpha) \text{BGE}$} \\
\midrule
0.1  & 35.6 & 80.5 & 90.0 & 40.7 & 80.7 & 90.3 \\
0.3  & 35.6 & 80.5 & 90.0 & \textbf{41.0} & \textbf{81.1} & \textbf{90.3} \\
0.5  & \textbf{35.3} & \textbf{81.3} & \textbf{90.4} & 40.4 & 80.9 & 90.1 \\
0.7  & 34.2 & 80.7 & 90.1 & 38.8 & 79.6 & 90.0 \\
0.9  & 32.9 & 79.7 & 89.7 & 36.2 & 78.1 & 89.5 \\
\bottomrule
\end{tabular}
\caption{Hybrid hyperparameter comparison on General QA.}
\label{tab:alpha_generalqa}
\end{table}

\subsection{ROUGE-L Comparison}
\begin{table*}[hbt]

\centering
\resizebox{\textwidth}{!}{
\begin{tabular}{l|p{22cm}}
\toprule
\multicolumn{2}{c}{\textbf{e-CARE Task 1}} \\
\midrule
\textbf{Question} & Tom hardly drinks water. \\
\textbf{Passage} & He suffered from kidney stones.. \\
\textbf{ROUGE-L} & 0.0 \\

\midrule

\textbf{Question} & Why some plants do not produce pollen? \\
\textbf{Passage} & Because some plants do not have male sex cells.
\\
\textbf{ROUGE-L} & 0.5 \\

\midrule

\multicolumn{2}{c}{\textbf{MS MARCO (CausalQA)}} \\

\midrule

\textbf{Question} & Why do my breasts hurt after menopause? \\
\textbf{Passage} & However, in most cases, unbalanced levels of estrogen and progesterone are the main cause behind menopausal breast tenderness. \\
\textbf{ROUGE-L} & 0.083 \\

\midrule

\textbf{Question} & What are the causes of the bubonic plague? \\
\textbf{Passage} & The causes of bubonic plague are bacteria called \textit{Yersinia pestis}. \\
\textbf{ROUGE-L Score} & 0.563 \\

\bottomrule
\end{tabular}}
\caption{Dataset examples with varying ROUGE-L scores}
\label{tab:rouge_examples}
\end{table*}

\label{appendix_rouge}
We report only the bins with more than 40 samples in the main text.  
Examples for each ROUGE score are shown in Table~\ref{tab:rouge_examples}.  
The full bin-wise sample counts for each dataset are listed below (Table~\ref{appendix:rouge_bins_e_care}, Table~\ref{appendix:rouge_bins_msmarco}).

\begin{table}[!h]
\centering
\caption{Number of samples per ROUGE score bin in e-CARE dataset.}
\label{appendix:rouge_bins_e_care} 
\textbf{E-CARE}
\vspace{0.5em}

\begin{tabular}{c|r}
\toprule
\textbf{ROUGE Bin} & \textbf{Count} \\
\midrule
{[}0.0, 0.1) & 749 \\
{[}0.1, 0.2) & 630 \\
{[}0.2, 0.3) & 421 \\
{[}0.3, 0.4) & 189 \\
{[}0.4, 0.5) & 87 \\
{[}0.5, 0.6) & 44 \\
{[}0.6, 0.7) & 6 \\
{[}0.7, 0.8) & 6 \\
{[}0.8, 0.9) & 2 \\
{[}0.9, 1.0) & 2 \\
\bottomrule
\end{tabular}
\end{table}

\begin{table}[h]
\centering
\caption{Number of samples per ROUGE score bin in MS MARCO (CausalQA) dataset.}
\label{appendix:rouge_bins_msmarco} 
\textbf{MS MARCO (CausalQA)}
\vspace{0.5em}

\begin{tabular}{c|r}
\toprule
\textbf{ROUGE Bin} & \textbf{Count} \\
\midrule
{[}0.00, 0.05) & 311 \\
{[}0.05, 0.10) & 379 \\
{[}0.10, 0.15) & 446 \\
{[}0.15, 0.20) & 433 \\
{[}0.20, 0.25) & 293 \\
{[}0.25, 0.30) & 229 \\
{[}0.30, 0.35) & 120 \\
{[}0.35, 0.40) & 79 \\
{[}0.40, 0.45) & 49 \\
{[}0.45, 0.50) & 18 \\
{[}0.50, 0.55) & 25 \\
{[}0.55, 0.60) & 12 \\
{[}0.60, 0.65) & 5 \\
{[}0.65, 0.70) & 5 \\
{[}0.70, 0.75) & 3 \\
{[}0.75, 0.80) & 0 \\
{[}0.80, 0.85) & 1 \\
{[}0.85, 0.90) & 2 \\
{[}0.90, 0.95) & 1 \\
{[}0.95, 1.00) & 3 \\
\bottomrule
\end{tabular}
\end{table}

\subsection{Relation between Cosine Similarity and Pearson Correlation}
\label{proof_corr}

The cosine similarity between two vectors \( X \) and \( Y \) is defined as

\[
\cos(X, Y) = \frac{X \cdot Y}{\|X\| \|Y\|}
\]
Where:
\[
X \cdot Y = \sum_{i=1}^n x_i y_i \quad
\]

\[
\|X\| = \sqrt{\sum_{i=1}^n x_i^2}, \quad \|Y\| = \sqrt{\sum_{i=1}^n y_i^2}  
\]

The Pearson correlation coefficient between two vectors \( X \) and \( Y \) is defined as

\[
\text{corr}(X, Y) = \frac{\sum_{i=1}^n (x_i - \bar{x})(y_i - \bar{y})}{\sqrt{\sum_{i=1}^n (x_i - \bar{x})^2} \sqrt{\sum_{i=1}^n (y_i - \bar{y})^2}}
\]

where

\[
\bar{x} = \frac{1}{n} \sum_{i=1}^n x_i, \quad  \bar{y} = \frac{1}{n} \sum_{i=1}^n y_i
\]

From this we can see that

\[
\text{corr}(X, Y) = \cos(X - \bar{X}, Y - \bar{Y})
\]

\bibliographystyle{IEEEtran}\bibliography{ref}

\begin{IEEEbiography}[{\includegraphics[width=1in,height=1.25in,clip,keepaspectratio]{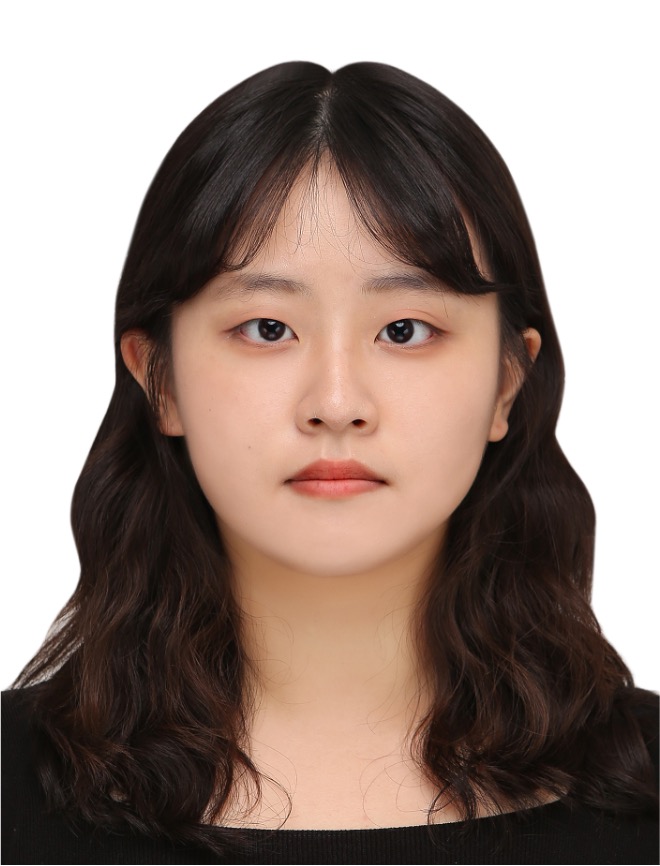}}]{Hyunseo Shin} is currently pursuing the M.S. degree in artificial intelligence at the University of Seoul, Seoul, Korea. Her research interests include retrieval methods, model merging, and natural language processing (NLP).
\end{IEEEbiography}

\begin{IEEEbiography}[{\includegraphics[width=1in,height=1.25in,clip,keepaspectratio]{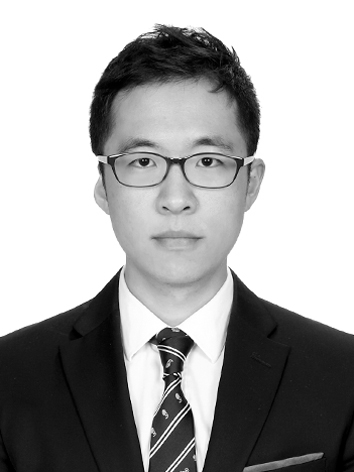}}]{Young Rok Choi}  received the B.A. degree in psychology and the B.S. degree in computer engineering from Sungkyunkwan University (SKKU), Suwon, South Korea, in 2015, and the M.Sc. degree in engineering from the Department of Applied Data Science, SKKU, in 2019. He is currently working at Naver Cloud, Inc., Seongnam, South Korea. His research interests include AI, NLP and information retrieval.

\end{IEEEbiography}

\begin{IEEEbiography}[{\includegraphics[width=1in,height=1.25in,clip,keepaspectratio]{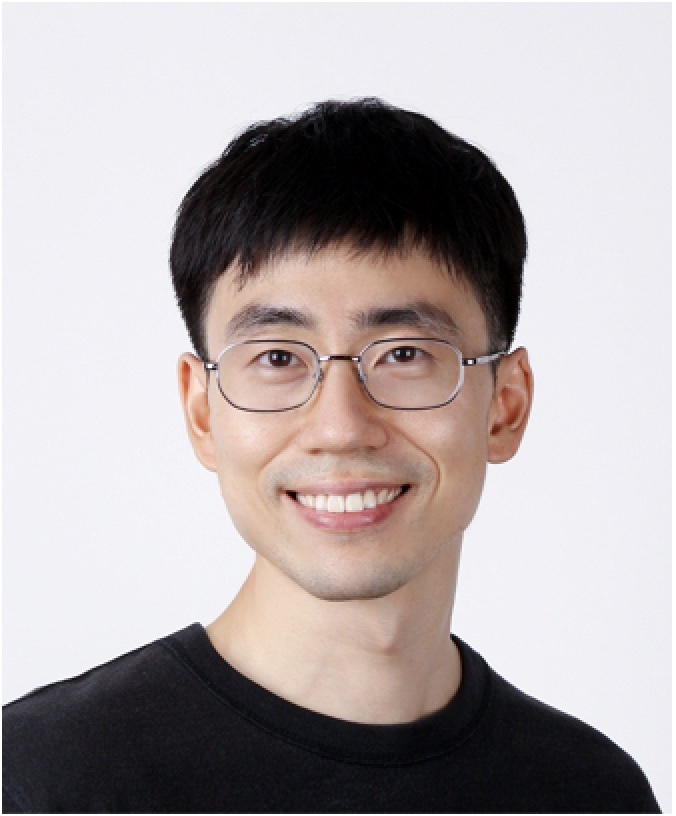}}]{Wonseok Hwang} is an Assistant Professor of Artificial Intelligence at the University of Seoul. His research areas include information retrieval, information extraction, AI alignment, and NLP application in legal and bio domains.
\end{IEEEbiography}

\EOD

\end{document}